\documentclass[10pt,twocolumn,letterpaper]{article}

\usepackage{cvpr}
\usepackage{times}
\usepackage{epsfig}
\usepackage{graphicx}
\usepackage{amsmath}
\usepackage{amssymb}
\usepackage{multirow}
\usepackage{latexsym}

\usepackage[pagebackref=true,breaklinks=true,letterpaper=true,colorlinks,bookmarks=false]{hyperref}

\cvprfinalcopy 

\def\cvprPaperID{643} 

\newcommand{\specialcell}[2][c]{%
	\begin{tabular}[#1]{@{}c@{}}#2\end{tabular}}
\begin{document}

\title{Cascaded Pyramid Network for Multi-Person Pose Estimation}

\author{
    Yilun Chen\thanks{:The first two authors contribute equally to this work. This work is done when Yilun Chen, Xiangyu Peng and Zhiqiang Zhang are interns at Megvii Research.} \quad Zhicheng Wang$^{*}$ \quad Yuxiang Peng$^{1}$ \quad Zhiqiang Zhang$^{2}$ \quad Gang Yu \quad Jian Sun \vspace{0.10cm} \\
    $^1$Tsinghua University   
    $^2$HuaZhong University of Science and Technology \\
    Megvii Inc. (Face++), \{chenyilun, wangzhicheng, pyx, zhangzhiqiang, yugang, sunjian\}@megvii.com 
}

\maketitle

\begin{abstract}
     The topic of multi-person pose estimation has been largely improved recently, especially with the development of convolutional neural network. However, there still exist a lot of challenging cases, such as occluded keypoints, invisible keypoints and complex background, which cannot be well addressed. In this paper, we present a novel network structure called Cascaded Pyramid Network~(CPN) which targets to relieve the problem from these ``hard'' keypoints. More specifically, our algorithm includes two stages: GlobalNet and RefineNet. GlobalNet is a feature pyramid network which can successfully localize the ``simple'' keypoints like eyes and hands but may fail to precisely recognize the occluded or invisible keypoints. Our RefineNet tries explicitly handling the ``hard'' keypoints by integrating all levels of feature representations from the GlobalNet together with an online hard keypoint mining loss. In general, to address the multi-person pose estimation problem, a top-down pipeline is adopted to first generate a set of human bounding boxes based on a detector, followed by our CPN for keypoint localization in each human bounding box. Based on the proposed algorithm, we achieve state-of-art results on the COCO keypoint benchmark, with average precision at 73.0 on the COCO test-dev dataset and 72.1 on the COCO test-challenge dataset, which is a 19\% relative improvement compared with 60.5 from the COCO 2016 keypoint challenge. Code\footnote{\color{red}\url{https://github.com/chenyilun95/tf-cpn.git}} and the detection results are publicly available for further research.
\end{abstract}

\section{Introduction}
    
    Multi-person pose estimation is to recognize and locate the keypoints for all persons in the image, which is a fundamental research topic for many visual applications like human action recognition and human-computer interaction. 
    
    
    Recently, the problem of multi-person pose estimation has been greatly improved by the involvement of deep convolutional neural networks~\cite{L1998Gradient,He2015}. For example, in~\cite{cao2017realtime}, convolutional pose machine is utilized to locate the keypoint joints in the image and part affinity fields~(PAFs) is proposed to assemble the joints to different person. Mask-RCNN~\cite{he2017maskrcnn} predicts human bounding boxes first and then warps the feature maps based on the human bounding boxes to obtain human keypoints. Although great progress has been made, there still exist a lot of challenging cases, such as occluded keypoints, invisible keypoints and crowded background, which cannot be well localized. The main reasons lie at two points: 1) these ``hard'' joints cannot be simply recognized based on their appearance features only, for example, the torso point; 2) these ``hard'' joints are not explicitly addressed during the training process.

    To address these ``hard'' joints, in this paper, we propose a novel network structure called Cascaded Pyramid Network~(CPN). There are two stages in our network architecture: GlobalNet and RefineNet. Our GlobalNet learns a good feature representation based on feature pyramid network~\cite{lin2016fpn}. More importantly, the pyramid feature representation can provide sufficient context information, which is inevitable for the inference of the occluded and invisible joints. Based on the pyramid features, our RefineNet explicitly address the ``hard'' joints based on an online hard keypoints mining loss.

    Based on our Cascaded Pyramid Network, we address the multi-person pose estimation problem based on a top-down pipeline. Human detector is first adopted to generate a set of human bounding boxes, followed by our CPN for keypoint localization in each human bounding box. In addition, we also explore the effects of various factors which might affect the performance of multi-person pose estimation, including person detector and data preprocessing. These details are valuable for the further improvement of accuracy and robustness of our algorithm.

    In summary, our contributions are three-fold as follows:
    \begin{itemize}
        \item We propose a novel and effective network called cascaded pyramid network~(CPN), which integrates global pyramid network~(GlobalNet) and pyramid refined network based on online hard keypoints mining~(RefineNet)
        \item We explore the effects of various factors contributing to muti-person pose estimation involved in top-down pipeline.
        \item Our algorithm achieves state-of-art results in the challenging COCO multi-person keypoint benchmark, that is, 73.0 AP in test-dev dataset and 72.1 AP in test challenge dataset.
    \end{itemize}

\section{Related Work}
    Human pose estimation is an active research topic for decades. Classical approaches tackling the problem of human pose estimation mainly adopt the techniques of pictorial structures~\cite{Fischler2006The,Andriluka2009Pictorial} or graphical models~\cite{Chen2014Articulated}. More specifically, the classical works~\cite{Andriluka2009Pictorial,classic_5,classic_6,classic_0,classic_2,classic_1,classic_4,classic_3} formulate the problem of human keypoints estimation as a tree-structured or graphical model problem and predict keypoint locations based on hand-crafted features. Recent works~\cite{Newell2016Stacked,Gkioxari2016Chained,nn_2_Bulat2016Human,nn_3_Insafutdinov2016DeeperCut,cpm,Yang_2017_ICCV} mostly rely on the development of convolutional neural network(CNN)~\cite{L1998Gradient,He2015}, which largely improve the performance of pose estimation. In this paper, we mainly focus on the methods based on the convolutional neural network. The topic is categorized as single-person pose estimation that predicts the human keypoints based on the cropped image given bounding box, and multi-person pose estimation that require further recognition of the full body poses of all persons in one image.

    {\bfseries Multi-Person Pose Estimation.}
        Multi-person pose estimation is gaining increasing popularity recently because of the high demand for the real-life applications. However, multi-person pose estimation is challenging owing to occlusion, various gestures of individual persons and unpredictable interactions between different persons. The approach of multi-person pose estimation is mainly divided into two categories: bottom-up approaches and top-down approaches.
        \begin{figure*}
            \begin{center}
                \includegraphics[width=150mm]{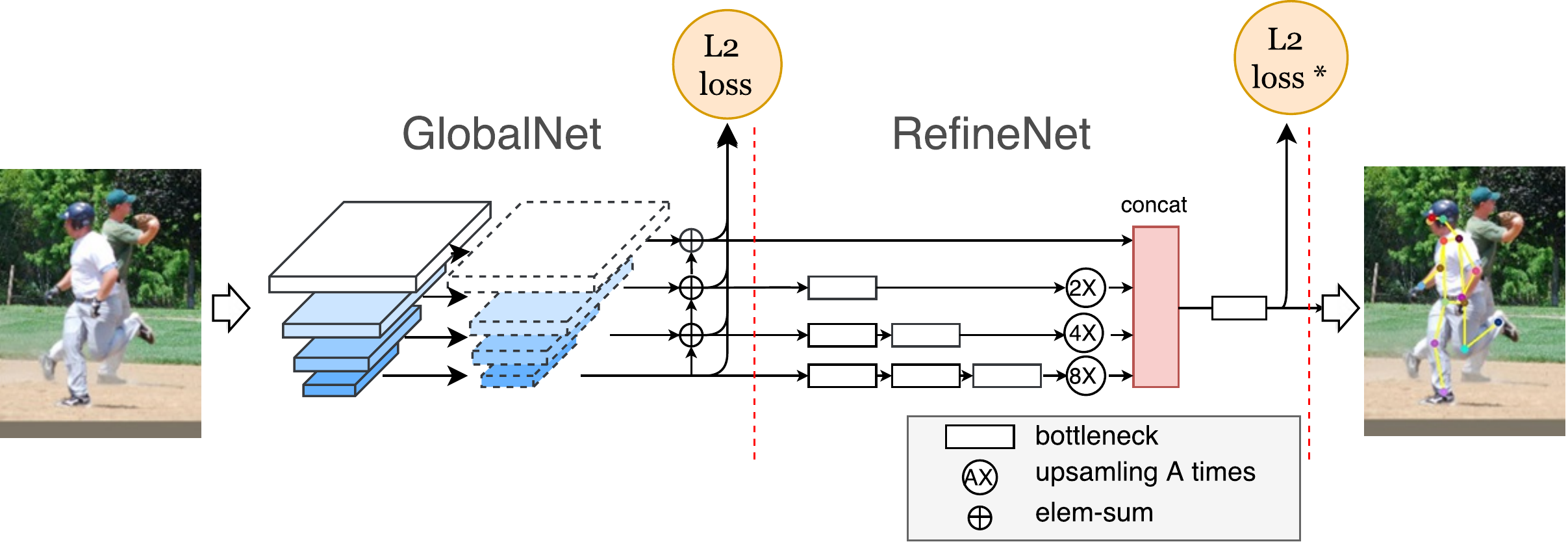}
            \end{center}
            \caption{Cascaded Pyramid Network. ``L2 loss*'' means L2 loss with online hard keypoints mining.}
            \label{fig:CPN}
        \end{figure*}

    {\bfseries Bottom-Up Approaches.}
        Bottom-up approaches~\cite{cao2017realtime,2016arXiv161105424N,Pishchulin2016DeepCut,nn_3_Insafutdinov2016DeeperCut} directly predict all keypoints at first and assemble them into full poses of all persons. DeepCut~\cite{Pishchulin2016DeepCut} interprets the problem of distinguishing different persons in an image as an Integer Linear Program (ILP) problem and partition part detection candidates into person clusters. Then the final pose estimation results are obtained when person clusters are combined with labeled body parts. DeeperCut~\cite{nn_3_Insafutdinov2016DeeperCut} improves DeepCut~\cite{Pishchulin2016DeepCut} using deeper ResNet~\cite{He2015} and employs image-conditioned pairwise terms to get better performance. Zhe Cao \etal.~\cite{cao2017realtime} map the relationship between keypoints into part affinity fields~(PAFs) and assemble detected keypoints into different poses of people. Newell \etal.~\cite{2016arXiv161105424N} simultaneously produce score maps and pixel-wise embedding to group the candidate keypoints to different people to get final multi-person pose estimation.
        
    {\bfseries Top-Down Approaches.}
        Top-down approaches~\cite{2017arXiv170101779P,Huang_2017_ICCV,he2017maskrcnn,RMPE} interpret the process of detecting keypoints as a two-stage pipeline, that is, firstly locate and crop all persons from image, and then solve the single person pose estimation problem in the cropped person patches. Papandreou \etal.~\cite{2017arXiv170101779P} predict both heatmaps and offsets of the points on the heatmaps to the ground truth location, and then uses the heatmaps with offsets to obtain the final predicted location of keypoints. Mask-RCNN~\cite{he2017maskrcnn} predicts human bounding boxes first and then crops the feature map of the corresponding human bounding box to predict human keypoints.
        If top-down approach is utilized for multi-person pose estimation, a human detector as well as single person pose estimator is important in order to obtain a good performance. Here we review some works about single person pose estimation and recent state-of-art detection methods.
        
    {\bfseries Single Person Pose Estimation.}
        Toshev \etal. firstly introduce CNN to solve pose estimation problem in the work of DeepPose~\cite{Toshev2013DeepPose}, which proposes a cascade of CNN pose regressors to deal with pose estimation. Tompson \etal.~\cite{Tompson2014Joint} attempt to solve the problem by predicting heatmaps of keypoints using CNN and graphical models. Later works such as Wei \etal.~\cite{Wei2016Convolutional}and Newell \etal.~\cite{Newell2016Stacked} show great performance via generating the score map of keypoints using very deep convolutional neural networks. Wei \etal.~\cite{Wei2016Convolutional} propose
        a multi-stage architecture, i.e., first generate coarse results, and continuously refine the result in the following stages. Newell \etal.~\cite{Newell2016Stacked} propose an U-shape network, i.e., hourglass module, and stack up several hourglass modules to generate prediction. Carreira \etal. \cite{Carreira2015Human} uses iterative error feedback to get pose estimation and refine the prediction gradually. Lifshitz \etal.~\cite{Lifshitz2016Human} uses deep consensus voting to vote the most probable location of keypoints. Gkioxary \etal.~\cite{Gkioxari2016Chained} and Zisserman \etal.~\cite{Belagiannis2016Recurrent} apply RNN-like architectures to sequentially refine the results. Our work is partly inspired by the works on generating  and refining score maps. Yang~\etal.~\cite{Feature_Pyramids} adopts pyramid features as inputs of the network in the process of pose estimation, which is a good exploration of the utilization of pyramid features in pose estimation. However, more refinement operations are required to pyramid structure in pose estimation.
        
    {\bfseries Human Detection.}
        Human detection approaches are mainly guided by the RCNN family~\cite{girshick2014rcnn,girshick15fastrcnn,ren2015faster}, the up-to-date detectors of which are ~\cite{lin2016fpn,he2017maskrcnn}. These detection approaches are composed of two-stage in general. First generate boxes proposals based on default anchors, and then crop from the feature map and further refine the proposals to get the final boxes via R-CNN network. The detector used in our methods are mostly based on \cite{lin2016fpn,he2017maskrcnn}.
\section{Our Approach for Multi-perosn Keypoints Estimation} \label{Network Architecture}



    Similar to~\cite{he2017maskrcnn,2017arXiv170101779P}, our algorithm adopts the top-down pipeline: a human detector is first applied on the image to generate a set of human bounding-boxes and detailed localization of the keypoints for each person can be predicted by a single-person skeleton estimator.

    \subsection{Human Detector}
    We adopt the state-of-art object detector algorithms based on FPN~\cite{lin2016fpn}. ROIAlign from Mask RCNN~\cite{he2017maskrcnn} is adopted to replace the ROIPooling in FPN. To train the object detector, all eighty categories from the COCO dataset are utilized during the training process but only the boxes of human category is used for our multi-person skeleton task. For fair comparison with our algorithms, we will release the detector results on the COCO val and COCO test dataset. 

        


        \begin{figure*}
            \begin{center}
                \includegraphics[width=150mm]{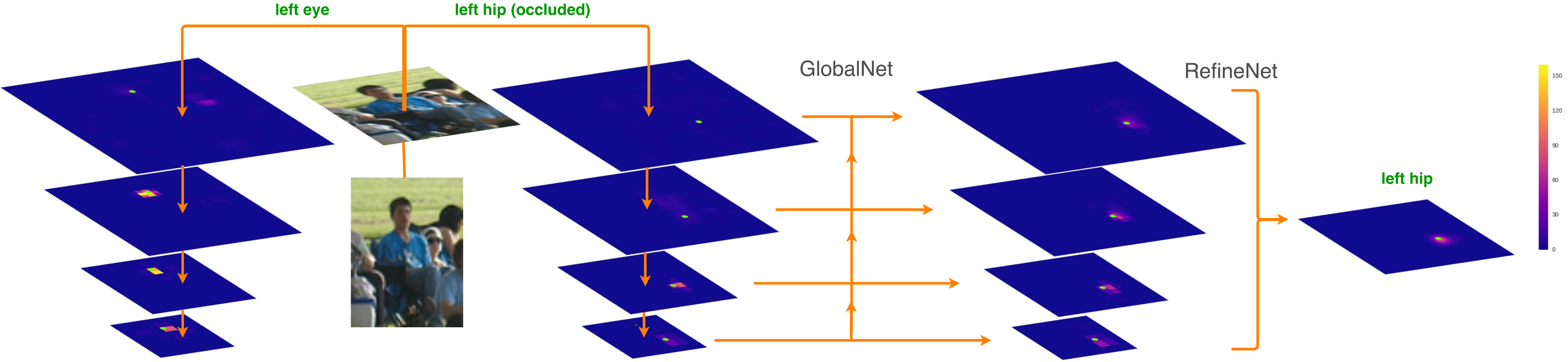}
            \end{center}
            \caption{Output heatmaps from different features. The green dots means the groundtruth location of keypoints.}
            \label{fig:heatmaps}
        \end{figure*}

    \subsection{Cascaded Pyramid Network (CPN)}

    Before starting the discussion of our CPN, we first briefly review the design structure for the single person pose estimator based on each human bounding box. Stacked hourglass~\cite{Newell2016Stacked}, which is a prevalent method for pose estimation, stacks eight hourglasses which are down-sampled and up-sampled modules with residual connections to enhance the pose estimation performance. The stacking strategy works to some extent, however, we find that stacking two hourglasses is sufficient to have a comparable performance compared with the eight-stage stacked hourglass module. \cite{2017arXiv170101779P} utilizes a ResNet~\cite{He2015} network to estimate pose in the wild achieving promising performance in the COCO 2016 keypoint challenge. Motivated by the works~\cite{Newell2016Stacked,2017arXiv170101779P} described above, we propose an effective and efficient network called cascaded pyramid network~(CPN) to address the problem of pose estimation. As shown in Figure~\ref{fig:CPN}, our CPN involves two sub-networks: GlobalNet and RefineNet. 
    \subsubsection{GlobalNet}\label{How network sees the keypoints?}

    Here, we describe our network structure based on the ResNet backbone. We denote the last residual blocks of different conv features conv2$\sim$5 as $C_2, C_3, ..., C_5$ respectively. $3\times 3$ convolution filters are applied on $C_2, ..., C_5$ to generate the heatmaps for keypoints. As shown in Figure~\ref{fig:heatmaps}, the shallow features like $C_2$ and $C_3$ have the high spatial resolution for localization but low semantic information for recognition. On the other hand, deep feature layers like $C_4$ and $C_5$ have more semantic information but low spatial resolution due to strided convolution (and pooling). Thus, usually an U-shape structure is integrated to maintain both the spatial resolution and semantic information for the feature layers. More recently, FPN~\cite{lin2016fpn} further improves the U-shape structure with deeply supervised information. We apply the similar feature pyramid structure for our keypoints estimation. Slightly different from FPN, we apply $1\times 1$ convolutional kernel before each element-wise sum procedure in the upsampling process. We call this structure as GlobalNet and an illustrative example can be found in Figure~\ref{fig:CPN}.

    As shown in Figure~\ref{fig:heatmaps}, our GlobalNet based on ResNet backbone can effectively locate the keypoints like eyes but may fail to precisely locate the position of hips. The localization of keypoints like hip usually requires more context information and processing rather than the nearby appearance feature. There exists many cases that are difficult to directly recognize these ``hard'' keypoints by a single GlobalNet. 
 
        
    \subsubsection{RefineNet}
    Based on the feature pyramid representation generated by GlobalNet, we attach a RefineNet to explicitly address the ``hard'' keypoints. In order to improve the efficiency and keep integrity of information transmission, our RefineNet transmits the information across different levels and finally integrates the informations of different levels via upsampling and concatenating as HyperNet~\cite{Kong2016HyperNet}. Different from the refinement strategy like stacked hourglass~\cite{Newell2016Stacked}, our RefineNet concatenates all the pyramid features rather than simply using the upsampled features at the end of hourglass module. In addition, we stack more bottleneck blocks into deeper layers, whose smaller spatial size achieves a good trade-off between effectiveness and efficiency.

    As the network continues training, the network tends to pay more attention to the ``simple'' keypoints of the majority but less importance to the occluded and hard keypoints. We should ensure the network balance between these two type of keypoints. Thus, in our RefineNet, we explictily select the hard keypoints online based on the training loss (which we called online hard keypoints mining) and back-propagate the gradients from the selected keypoints only.

\section{Experiment}\label{Experiment}
    
    Our overall pipeline follows the top-down approach for estimating multiple human poses. Firstly, we apply a state-of-art bounding detector to generate human proposals. For each proposal, we assume that there is only one main person in the cropped region of proposal and then applied the pose estimating network to generate the final prediction. In this section, we will discuss more details of our methods based on experiment results.

    \subsection{Experimental Setup}
    
    {\bfseries Dataset and Evaluation Metric.}
        Our models are only trained on MS COCO\cite{Lin2014Microsoft} trainval dataset (includes 57K images and 150K person instances) and validated on MS COCO minival dataset (includes 5000 images). The testing sets includes test-dev set (20K images) and test-challenge set~(20K images). Most experiments are evaluated in OKS-based mAP, where OKS~(object keypoints similarity) defines the similarity between different human poses. 

    {\bfseries Cropping Strategy.}
        For each human detection box, the box is extended to a fixed aspect ratio, e.g., height : width = 256 : 192, and then we crop from images without distorting the images aspect ratio. Finally, we resize the cropped image to a fixed size of height 256 pixels and 192 pixels by default. Note that only the boxes of the person class in the top 100 boxes of all classes are used in all the experiments of~\ref{ablation expriments}.

    {\bfseries Data Augmentation Strategy.}
        Data augmentation is critical for the learning of scale invariance and rotation invariance. After cropping from images, we apply random flip, random rotation ($-45^\circ\sim+45^\circ$) and random scale ($0.7\sim 1.35$). 

    {\bfseries Training Details.}
        All models of pose estimation are trained using adam algorithm with an initial learning rate of 5e-4. Note that we also decrease the learning rate by a factor of 2 every 3600000 iteration. We use a weight decay of 1e-5 and the training batch size is 32. Batch normalization is used in our network. Generally, the training of ResNet-50-based models takes about 1.5 day on eight NVIDIA Titan X Pascal GPUs. Our models are all initialized with weights of the public-released ImageNet~\cite{ILSVRC15}-pretrained model. 
        

    {\bfseries Testing Details.}
        In order to minimize the variance of prediction, we apply a gaussian filter on the predicted heatmaps. Following the same techniques used in \cite{Newell2016Stacked}, we also predict the pose of the corresponding flipped image and average the heatmaps to get the final prediction; a quarter offset in the direction from the highest response to the second highest response is used to obtain the final location of the keypoints. Rescoring strategy is also used in our experiments. Different from the rescoring strategy used in \cite{2017arXiv170101779P}, the product of boxes' score and the average score of all keypoints is considered as the final pose score of a person instance.
    
    \subsection{Ablation Experiment}\label{ablation expriments}
    In this subsection, we'll validate the effectiveness of our network from various aspects. Unless otherwise specified, all experiments are evaluated on MS COCO minival dataset in this subsection. The input size of all models is $256\times 192$ and the same data augmentation is adopted.
    
    \subsubsection{Person Detector}
    Since detection boxes are critical for top-down approaches in multi-person pose estimation, here we discuss two factors of detection, i.e. different NMS strategies and the AP of bounding boxes. Our human boxes are generated based on the state-of-art detector FPN trained with only the labeled COCO data, no extra data and no specific training on person. For fair comparison, we use the same detector with a general AP of 41.1 and person AP of 55.3 on the COCO minival dataset in the ablation experiments by default unless otherwise specified.
    
    {\bfseries Non-Maximum Suppression~(NMS) strategies.} 
    As shown in the Table~\ref{table:Relationship between NMS methods and keypoints AP}, we compare the performance of different NMS strategies or the same NMS strategy under different thresholds. Referring to the original hard NMS, the performance of keypoints detection improves when the threshold increases, basically owing to the improvement of the average precision~(AP) and average recall~(AR) of the boxes. Since the final score of the pose estimated partially depends on the score of the bounding box, Soft-NMS~\cite{softNMS} which is supposed to generate more proper scores is better in performance as it is shown in the Table~\ref{table:Relationship between NMS methods and keypoints AP} . From the table, we can see that Soft-NMS~\cite{softNMS} surpasses the hard NMS method on the performance of both detection and keypoints detection.
    \begin{table}[h]\small
        \begin{center}
            \begin{tabular}{|l|c|c|c|c|c|c|c|c|}
                \hline
                NMS &AP(all)&AP(H)&AR(H)&AP(OKS) \\
                \hline
                NMS(thr=0.3) & 40.1 & 53.5 & 60.3 & 68.2 \\ 
                \hline
                NMS(thr=0.4) & 40.5 & 54.4 & 61.7 & 68.9 \\ 
                \hline
                NMS(thr=0.5) & 40.8 & 54.9 & 62.9 & 69.2 \\ 
                \hline
                NMS(thr=0.6) & 40.8 & 55.2 & 64.3 & 69.2 \\ 
                \hline
                Soft-NMS~\cite{softNMS} & 41.1 & 55.3 & 67.0 & 69.4 \\ 
                \hline
            \end{tabular}
        \end{center}
        \caption{Comparison between different NMS methods and keypoints detection performance with the same model. H is short for human.}
        \label{table:Relationship between NMS methods and keypoints AP}
    \end{table}
    
    {\bf Detection Performance.} Table~\ref{table:Relationship between detection AP and keypoints AP} shows the relationship between detection AP and the corresponding keypoints AP, aiming to reveal the influence of the accuracy of the bounding box detection on the keyoints detection. From the table, we can see that the keypoints detection AP gains less and less as the accuracy of the detection boxes increases. Specially, when the detection AP increases from 44.3 to 49.3 and the human detection AP increases 3.0 points, the keypoints detection accuracy does not improve a bit and the AR of the detection increases marginally. Therefore, we have enough reasons to deem that the given boxes cover most of the medium and large person instances with such a high detection AP. Therefore, the more important problem for pose estimation is to enhance the accuracy of hard keypoints other than involve more boxes.
    \begin{table}[h]\small
        \begin{center}
            \begin{tabular}{|l|c|c|c|c|c|c|c|c|}
                \hline
                Det Methods & AP(all)&AP(H)&AR(H)&AP(OKS) \\
                \hline
                FPN-1 & 36.3 & 49.6 & 58.5 & 68.8 \\ \hline
                FPN-2 & 41.1 & 55.3 & 67.0 & 69.4 \\ \hline
                FPN-3 & 44.3 & 58.4 & 71.3 & 69.7 \\ \hline
                ensemble-1 & 49.3 & 61.4 & 71.8 & 69.8 \\ \hline
                ensemble-2 & 52.1 & 62.9 & 74.7 & 69.8 \\ \hline
            \end{tabular}
        \end{center}
        \caption{Comparison between detection performance and keypoints detection performance. FPN-1: FPN with the backbone of Res50; FPN-2: Res101 with Soft-NMS and OHEM~\cite{ohem} applied; FPN-3: ResNeXt~\cite{resnext}101 with Soft-NMS, OHEM~\cite{ohem}, multiscale training applied; ensemble-1: multiscale test involved; ensemble-2: multiscale test, large batch and SENet~\cite{Hu} involved. H is short for Human.}
        \label{table:Relationship between detection AP and keypoints AP}
    \end{table}

    \subsubsection{Cascaded Pyramid Network}\label{Visibility-Aware Network}
    8-stage hourglass network~\cite{Newell2016Stacked} and ResNet-50 with dilation~\cite{2017arXiv170101779P} are adopted as our baseline. From Table~\ref{table:Comparison on COCO minival dataset}, although the results improve considerably if dilation are used in shallow layers, it is worth noting that the FLOPs (floating-point operations) increases significantly. 
    
    \begin{table}[h]\small
        \begin{center}
            \begin{tabular}{|l|c|c|c|c|c|c|c|c|}
                \hline
                Models & AP~(OKS) & FLOPs & Param Size \\ \hline
                1-stage hourglass & 54.5 & 3.92G & 12MB \\ \hline
                2-stage hourglass & 66.5 & 6.14G & 23MB \\ \hline
                8-stage hourglass & 66.9 & 19.48G & 89MB \\ \hline
                ResNet-50 & 41.3 & 3.54G & 92MB \\ \hline
                \specialcell{ResNet-50 \ \ \ \ \ \\+ dilation(res5)} & 44.1 & 5.62G & 92MB \\ \hline
                \specialcell{ResNet-50\ \ \ \ \ \ \\ + dilation(res4-5)} & 66.5 & 17.71G & 92MB \\ \hline
                \specialcell{ResNet-50\ \ \ \ \ \ \\ + dilation(res3-5)} & -- & 68.70G & 92MB \\ \hline
                \specialcell{GlobalNet only\\(ResNet-50)} & 66.6 & 3.90G & 94MB \\ \hline
                \specialcell{CPN* (ResNet-50) }& 68.6 & 6.20G & 102 MB \\ \hline
                CPN (ResNet-50) & 69.4 & 6.20G & 102 MB \\ 
                \hline
            \end{tabular}
        \end{center}
        \caption{Results on COCO minival dataset. CPN* indicates CPN without online hard keypoints mining.}
        \label{table:Comparison on COCO minival dataset}
    \end{table}
    From the statistics of FLOPs in testing stage and the accuracy of keypoints as shown in Table~\ref{table:Comparison on COCO minival dataset}, we find that CPN achieves much better speed-accuracy trade-off than Hourglass network and ResNet-50 with dilation. Note that GlobalNet achieves much better results than one-stage hourglass network of same FLOPs probably for much larger parameter space. After refined by the RefineNet, it increases 2.0 AP and yields the results of 68.6 AP. Furthermore, when online hard keypoints mining is applied in RefineNet, our network finally achieves 69.4 AP.
    
    {\bfseries Design Choices of RefineNet.}
    Here, we compare different design strategies of RefineNet as shown in Table~\ref{table: refinenet design strategies}. 
    We compare the following implementation based on pyramid output from the GlobalNet: 
    \begin{itemize}
    \item[1)] Concatenate (Concat) operation is directly attached like HyperNet~\cite{Kong2016HyperNet},
    \item[2)] Only one bottleneck block is attached first in each layer ($C_2 \sim C_5$) and then followed by a concatenate operation,  
    \item[3)] Different number of bottleneck blocks applied to different layers followed by a concatenate operation as shown in Figure~\ref{fig:CPN}. 
    \end{itemize}
    A convolution layer is attached finally to generate the score maps for each keypoint. 
    
    We find that our RefineNet structure can effectively achieve more than 2 points gain compared with GlobalNet only and  for refinement of keypoints and also outperforms other design implementations followed by GlobalNet.
    \begin{table}[h]\small
        \begin{center}
            \begin{tabular}{|l|c|c|c|c|c|c|c|c|}
                \hline
                Models & AP(OKS) & FLOPs \\ \hline
                GlobalNet only & 66.6 & 3.90G \\ \hline
                GlobalNet + Concat & 68.5 & 5.87G \\ \hline
                GlobalNet + 1 bottleneck +Concat & 69.2 & 6.92G \\ \hline
                ours (CPN) & 69.4 & 6.20G \\ \hline
            \end{tabular}
        \end{center}
        \caption{Comparison of models of different design choices of RefineNet.}
        \label{table: refinenet design strategies}
    \end{table}
    
    Here, we also validate the performance for utilizing the pyramid output from different levels. In our RefineNet, we utilize four output feature maps $C_2\sim C_5$, where $C_i$ refers to the $i$th feature map of GlobalNet output. Also, feature map from $C_2$ only, feature maps from $C_2\sim C_3$, and feature maps from $C_2\sim C_4$ are evaluated as shown in Table~\ref{table: Effectiveness of RefineNet}. We can find that the performance improves as more levels of features are utilized. 
 
    \begin{table}[h]\small
        \begin{center}
            \begin{tabular}{|c|c|c|c|c|c|c|c|c|}
                \hline
                Connections & AP(OKS) & FLOPs \\ \hline
                $C_2$ & 68.3 & 5.02G \\ \hline
                $C_2\sim C_3$ & 68.4 & 5.50G \\ \hline
                $C_2\sim C_4$ & 69.1 & 5.88G \\ \hline
                $C_2\sim C_5$ & 69.4 & 6.20G \\ \hline
            \end{tabular}
        \end{center}
        \caption{Effectiveness of intermediate connections between GlobalNet and RefineNet.}
        \label{table: Effectiveness of RefineNet}
    \end{table}
    
    \subsubsection{Online Hard Keypoints Mining}
    Here we discuss the losses used in our network. In detail, the loss function of GlobalNet is L2 loss of all annotated keypoints while the second stage tries learning the hard keypoints, that is, we only punish the top $M (M < N)$ keypoint losses out of $N$ (the number of annotated keypoints in one person, say 17 in COCO dataset). The effect of $M$ is shown in Table~\ref{table: Comparison of different hard keypoints number in hard keypoint mining}. For $M = 8$, the performance of second stage achieves the best result for the balanced training between hard keypoints and simple keypoints.
    \begin{table}[h]\small
        \begin{center}
            \begin{tabular}{|c|c|c|c|c|c|c|c|c|}
                \hline
                $M$ & 6 & 8 & 10 & 12 & 14 & 17 \\ \hline
                AP~(OKS) & 68.8 & 69.4 & 69.0 & 69.0 & 69.0 & 68.6\\ \hline
            \end{tabular}
        \end{center}
        \caption{Comparison of different hard keypoints number in online hard keypoints mining.}
        \label{table: Comparison of different hard keypoints number in hard keypoint mining}
    \end{table}
    
    Inspired by OHEM~\cite{ohem}, however the method of online hard keypoints mining loss is essentially different from it. Our method focuses on higher level information than OHEM which concentrates on examples, for instance, pixel level losses in the heatmap L2 loss. As a result, our method is more stable, and outperforms OHEM strategy in accuracy. 
    
    As Table~\ref{table: visibility aware loss} shows, when online hard keypoints mining is applied in RefineNet, the performance of overall network increases 0.8 AP and finally achieves 69.4 AP comparing to normal l2 loss. For reference, experiments without intermediate supervision in CPN leads to a performance drop of 0.9 AP probably for the lack of prior knowledge and sufficient context information of keypoints provided by GlobalNet. In addition, applying the same online hard keypoints mining in GlobalNet which decreases the results by 0.3 AP.
    \begin{table}[h]\small
        \begin{center}
            \begin{tabular}{|c|c|c|c|c|c|c|c|c|}
                \hline
                GlobalNet & RefineNet & AP(OKS) \\ \hline
                --- & L2 loss & 68.2 \\  \hline
                L2 loss & L2 loss & 68.6 \\ \hline
                --- & L2 loss* & 68.5 \\ \hline
                L2 loss & L2 loss* & 69.4 \\ \hline
                L2 loss* & L2 loss* & 69.1 \\ \hline
            \end{tabular}
        \end{center}
        \caption{Comparison of models with different losses function. Here ``-'' denotes that the model applies no loss function in corresponding subnetwork. ``L2 loss*'' means L2 loss with online hard keypoints mining.}
        \label{table: visibility aware loss}
    \end{table}
    
    \subsubsection{Data Pre-processing}
    The size of cropped image are important factors to the performance of keypoints detection. As Table~\ref{table: cropping strategy} illustrates, it's worth noting that the input size $256\times 192$ actually works as well as $256\times 256$ which costs more computations of almost 2G FLOPs using the same cropping strategy. As the input size of the cropped images increases, more location details of human keypoints are fed into the network resulting in a large performance improvement. Additionally, online hard keypoints mining works better when the input size of the crop images is enlarged by improving 1 point on $384\times 288$ input size. 
    
    \begin{table}[h]\small
        \begin{center}
            \begin{tabular}{|l|c|c|c|c|c|c|c|c|}
                \hline
                Models & Input Size & FLOPs & AP(OKS) \\ \hline
                8-stage Hourglass & $256\times 192$ & 19.5G & 66.9 \\ \hline
                8-stage Hourglass & $256\times 256$ & 25.9G & 67.1  \\ \hline
                \specialcell{CPN* (ResNet-50)} & $256\times 192$ & 6.2G & 68.6 \\ \hline
                CPN (ResNet-50) & $256\times 192$ & 6.2G & 69.4 \\ \hline
                \specialcell{CPN* (ResNet-50)} & $384\times 288$ & 13.9G & 70.6  \\  \hline
                CPN (ResNet-50)& $384\times 288$ & 13.9G & 71.6\\
                \hline
            \end{tabular}
        \end{center}
        \caption{Comparison of models of different input size. CPN* indicates CPN without online hard keypoints mining.}
        \label{table: cropping strategy}
    \end{table}
    
    \begin{table*}[h]
        \begin{center}\small
            \begin{tabular}{|l|c|c|c|c|c|c|c|c|c|c|c|c|}
                \hline
                Methods  & AP &  $\text{AP}_{@.5}$ & $\text{AP}_{@.75}$ & $\text{AP}_m$ & $\text{AP}_l$ & \text{AR} & $\text{AR}_{@.5}$ & $\text{AR}_{@.75}$ & $\text{AR}_m$ & $\text{AR}_l$ \\ 
                \hline
                FAIR Mask R-CNN* & 68.9 & 89.2 & 75.2 & 63.7 & 76.8 & 75.4 & 93.2 & 81.2 & 70.2 & 82.6 \\ 
                \hline
                G-RMI* & 69.1&	85.9&	75.2&	66.0&	74.5&	75.1&	90.7&	80.7&	69.7&	82.4 \\
                \hline
                bangbangren+* & 70.6 &	88.0 &	76.5 &	65.6 &	{\bf 79.2 }&	77.4&	93.6&	83.0&	71.8&{\bf 85.0 }\\ 
                \hline
                oks* & 71.4 &	89.4&	78.1&	65.9&	79.1&	77.2&	93.6&	83.4&	71.8&	84.5 \\ 
                \hline
                {\bf Ours+~(CPN+) } & {\bf 72.1}&	{\bf 90.5}&	{\bf 78.9} & {\bf 67.9}& 78.1 &	{\bf 78.7 }& {\bf 94.7 }& {\bf 84.8 }& {\bf 74.3 }&	84.7 \\ \hline
            \end{tabular}
        \end{center}
        \caption{Comparisons of final results on COCO test-challenge2017 dataset. ``*'' means that the method involves extra data for training. Specifically, FAIR Mask R-CNN involves distilling unlabeled data, oks uses AI-Challenger keypoints dataset, bangbangren and G-RMI use their internal data as extra data to enhance performance. ``+'' indicates results using ensembled models. The human detector of Ours+ is a detector that has an AP of 62.9 of human class on COCO minival dataset. CPN and CPN+ in this table all use the backbone of ResNet-Inception~\cite{resinc} framework.}
        \label{table: Results test-challenge2017 dataset}
    \end{table*}
    
    \begin{table*}[h]
        \begin{center}\small
            \begin{tabular}{|l|c|c|c|c|c|c|c|c|c|c|c|c|}
                \hline
                Methods  & AP &  $\text{AP}_{@.5}$ & $\text{AP}_{@.75}$ & $\text{AP}_m$ & $\text{AP}_l$ & \text{AR} & $\text{AR}_{@.5}$ & $\text{AR}_{@.75}$ & $\text{AR}_m$ & $\text{AR}_l$ \\ 
                \hline
                CMU-Pose~\cite{cao2017realtime} & 61.8 & 84.9 & 67.5 & 57.1 & 68.2 & 66.5 & 87.2 & 71.8 & 60.6 & 74.6 \\ 
                \hline
                Mask-RCNN~\cite{he2017maskrcnn} & 63.1&	87.3& 68.7&	57.8& 71.4&	-&	-&	-&	-&	- \\
                \hline
                Associative Embedding~\cite{2016arXiv161105424N} & 65.5 & 86.8 & 72.3 &	60.6 & 72.6& 70.2&	89.5& 76.0&	64.6& 78.1 \\ 
                \hline
                G-RMI~\cite{2017arXiv170101779P} & 64.9 & 85.5& 71.3& 62.3& 70.0& 69.7& 88.7& 75.5& 64.4& 77.1 \\
                \hline
                G-RMI*~\cite{2017arXiv170101779P} & 68.5 & 87.1& 75.5& 65.8& 73.3& 73.3& 90.1& 79.5& 68.1&	80.4 \\ 
                \hline
                {\bf Ours~(CPN)} & 72.1 & 91.4&	80.0&	68.7&	77.2&	78.5&	{\bf 95.1}&	85.3&	74.2&	84.3 \\ 
                \hline
                {\bf Ours+~(CPN+)} & {\bf 73.0}&	{\bf 91.7}&	{\bf 80.9}&	{\bf 69.5}&	{\bf 78.1}&	{\bf 79.0}&	{\bf 95.1}&	{\bf 85.9}&	{\bf 74.8}&	{\bf 84.7} \\ \hline
            \end{tabular}
        \end{center}
        \caption{Comparisons of final results on COCO test-dev dataset. ``*'' means that the method involves extra data for training. ``+'' indicates results using ensembled models. The human detectors of Our and Ours+ the same detector that has an AP of 62.9 of human class on COCO minival dataset.CPN and CPN+ in this table all use the backbone of ResNet-Inception~\cite{resinc} framework.}
        \label{table: Results test-dev dataset}
    \end{table*}
    
    \begin{table}[h]\small
        \begin{center}
            \begin{tabular}{|l|c|c|c|c|c|c|c|c|}
                \hline
                Methods & AP - minival & AP - dev & AP - challenge \\ \hline
                Ours (CPN) & 72.7 & 72.1 & -  \\ \hline
                Ours (CPN+) & 74.5 & 73.0 & 72.1 \\ \hline
            \end{tabular}
        \end{center}
        \caption{Comparison of results on the minvival dataset and the corresponding results on test-dev or test-challenge of the COCO dataset. ``+'' indicates ensembled model. CPN and CPN+ in this table all use the backbone of ResNet-Inception~\cite{resinc} framework.}
        \label{table: Results on COCO minival dataset}
    \end{table}

    \subsection{Results on MS COCO Keypoints Challenge}
\begin{figure*}
    \begin{center}
        \includegraphics[width=152mm]{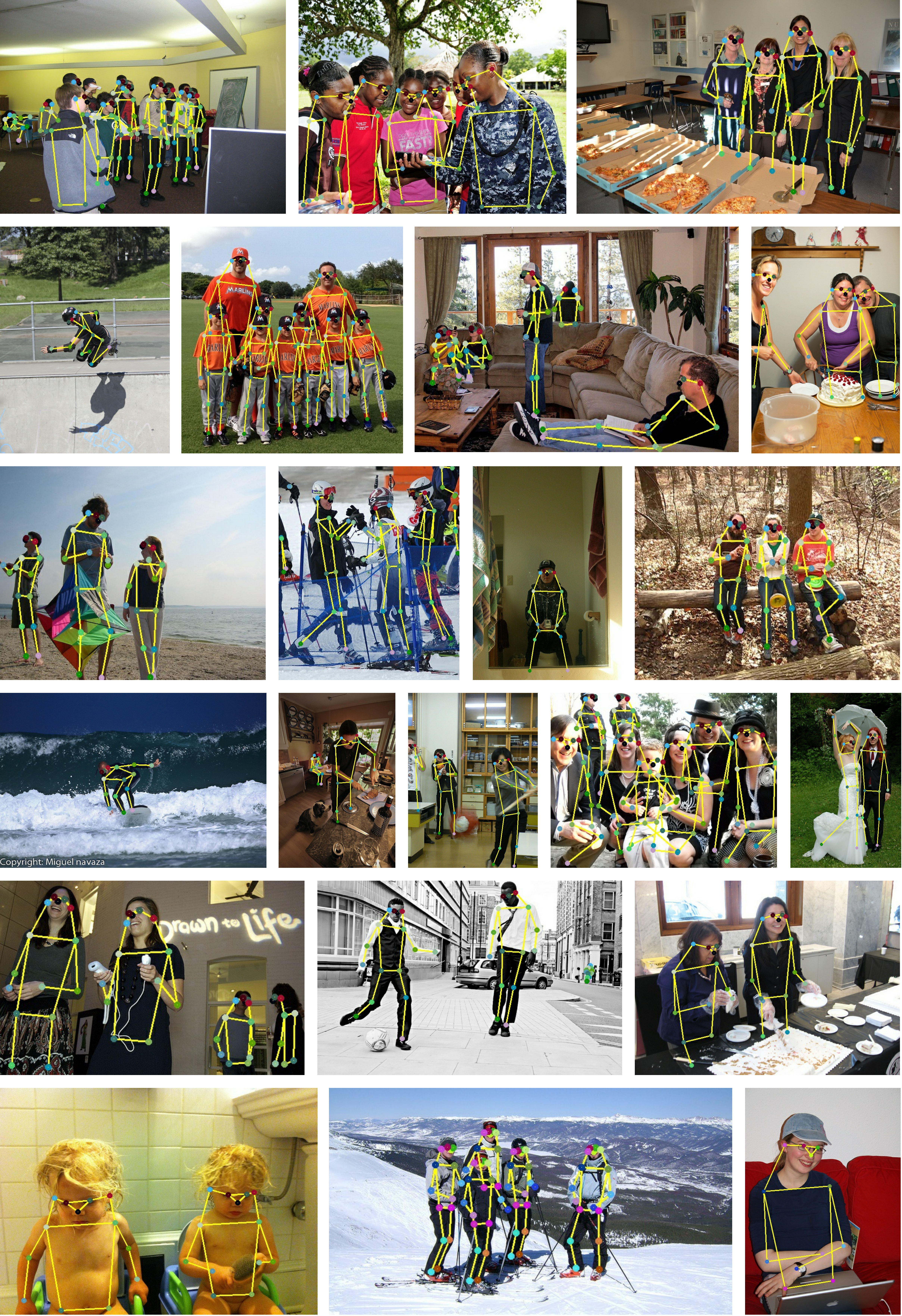}
    \end{center}
    \caption{Some results of our method.}
    \label{fig:illustrativeResults}
\end{figure*}

    We evaluate our method on MS COCO test-dev and test-challenge dataset. Table~\ref{table: Results test-dev dataset} illustrates the results of our method in the test-dev split dataset of the COCO dataset. Without extra data involved in training, we achieve 72.1 AP using a single model of CPN and 73.0 using ensembled models of CPN with different ground truth heatmaps. Table~\ref{table: Results test-challenge2017 dataset} shows the comparison of the results of our method and the other methods on the test-challenge2017 split of COCO dataset. We get 72.1 AP achieving state-of-art performance on COCO test-challenge2017 dataset. Table~\ref{table: Results on COCO minival dataset} shows the performances of CPN and CPN+ (ensembled model) on COCO minival dataset, which offer a reference to the gap between the COCO minival dataset and the standard test-dev or test-challenge dataset of the COCO dataset. Figure~\ref{fig:illustrativeResults} illustrates some results generated using our method. 
    
\section{Conclusion}
    In this paper, we follow the top-down pipeline and a novel Cascaded Pyramid Network~(CPN) is presented to address the ``hard'' keypoints. More specifically, our CPN includes a GlobalNet based on the feature pyramid structure and a RefineNet which concatenates all the pyramid features as a context information. In addition, online hard keypoint mining is integrated in RefineNet to explicitly address the ``hard'' keypoints. Our algorithm achieves state-of-art results on the COCO keypoint benchmark, with average precision at 73.0 on the COCO test-dev dataset and 72.1 on the COCO test-challenge dataset, outperforms the COCO 2016 keypoint challenge winner by a 19\% relative improvement. 

{\small
\bibliographystyle{ieee}
\bibliography{egbib}

\begin{thebibliography}{10}\itemsep=-1pt

\bibitem{Andriluka2009Pictorial}
M.~Andriluka, S.~Roth, and B.~Schiele.
\newblock Pictorial structures revisited: People detection and articulated pose
  estimation.
\newblock In {\em Computer Vision and Pattern Recognition, 2009. CVPR 2009.
  IEEE Conference on}, pages 1014--1021, 2009.

\bibitem{Belagiannis2016Recurrent}
V.~Belagiannis and A.~Zisserman.
\newblock Recurrent human pose estimation.
\newblock pages 468--475, 2016.

\bibitem{softNMS}
N.~{Bodla}, B.~{Singh}, R.~{Chellappa}, and L.~S. {Davis}.
\newblock {Improving Object Detection With One Line of Code}.
\newblock {\em ArXiv e-prints}, Apr. 2017.

\bibitem{nn_2_Bulat2016Human}
A.~Bulat and G.~Tzimiropoulos.
\newblock {\em Human Pose Estimation via Convolutional Part Heatmap
  Regression}.
\newblock Springer International Publishing, 2016.

\bibitem{cao2017realtime}
Z.~Cao, T.~Simon, S.-E. Wei, and Y.~Sheikh.
\newblock Realtime multi-person 2d pose estimation using part affinity fields.
\newblock In {\em CVPR}, 2017.

\bibitem{Carreira2015Human}
J.~Carreira, P.~Agrawal, K.~Fragkiadaki, and J.~Malik.
\newblock Human pose estimation with iterative error feedback.
\newblock 2013(2013):4733--4742, 2015.

\bibitem{Chen2014Articulated}
X.~Chen and A.~Yuille.
\newblock Articulated pose estimation by a graphical model with image dependent
  pairwise relations.
\newblock {\em Eprint Arxiv}, pages 1736--1744, 2014.

\bibitem{classic_2}
M.~Dantone, J.~Gall, C.~Leistner, and L.~Van~Gool.
\newblock Human pose estimation using body parts dependent joint regressors.
\newblock In {\em The IEEE Conference on Computer Vision and Pattern
  Recognition (CVPR)}, June 2013.

\bibitem{RMPE}
H.-S. Fang, S.~Xie, Y.-W. Tai, and C.~Lu.
\newblock Rmpe: Regional multi-person pose estimation.
\newblock In {\em The IEEE International Conference on Computer Vision (ICCV)},
  Oct 2017.

\bibitem{Fischler2006The}
M.~A. Fischler and R.~A. Elschlager.
\newblock The representation and matching of pictorial structures.
\newblock {\em IEEE Transactions on Computers}, C-22(1):67--92, 2006.

\bibitem{girshick15fastrcnn}
R.~Girshick.
\newblock Fast {R-CNN}.
\newblock In {\em Proceedings of the International Conference on Computer
  Vision ({ICCV})}, 2015.

\bibitem{girshick2014rcnn}
R.~Girshick, J.~Donahue, T.~Darrell, and J.~Malik.
\newblock Rich feature hierarchies for accurate object detection and semantic
  segmentation.
\newblock In {\em Proceedings of the IEEE Conference on Computer Vision and
  Pattern Recognition ({CVPR})}, 2014.

\bibitem{classic_6}
G.~Gkioxari, P.~Arbelaez, L.~Bourdev, and J.~Malik.
\newblock Articulated pose estimation using discriminative armlet classifiers.
\newblock In {\em IEEE Conference on Computer Vision and Pattern Recognition},
  pages 3342--3349, 2013.

\bibitem{Gkioxari2016Chained}
G.~Gkioxari, A.~Toshev, and N.~Jaitly.
\newblock Chained predictions using convolutional neural networks.
\newblock In {\em European Conference on Computer Vision}, pages 728--743,
  2016.

\bibitem{he2017maskrcnn}
K.~He, G.~Gkioxari, P.~Doll\'{a}r, and R.~Girshick.
\newblock {Mask R-CNN}.
\newblock {\em arXiv preprint arXiv:1703.06870}, 2017.

\bibitem{He2015}
K.~He, X.~Zhang, S.~Ren, and J.~Sun.
\newblock Deep residual learning for image recognition.
\newblock In {\em The IEEE Conference on Computer Vision and Pattern
  Recognition (CVPR)}, June 2016.

\bibitem{Hu}
J.~Hu, L.~Shen, and G.~Sun.
\newblock {Squeeze-and-Excitation Networks}.

\bibitem{Huang_2017_ICCV}
S.~Huang, M.~Gong, and D.~Tao.
\newblock A coarse-fine network for keypoint localization.
\newblock In {\em The IEEE International Conference on Computer Vision (ICCV)},
  Oct 2017.

\bibitem{nn_3_Insafutdinov2016DeeperCut}
E.~Insafutdinov, L.~Pishchulin, B.~Andres, M.~Andriluka, and B.~Schiele.
\newblock Deepercut: A deeper, stronger, and faster multi-person pose
  estimation model.
\newblock In {\em European Conference on Computer Vision}, pages 34--50, 2016.

\bibitem{classic_3}
S.~Johnson and M.~Everingham.
\newblock Learning effective human pose estimation from inaccurate annotation.
\newblock In {\em Computer Vision and Pattern Recognition}, pages 1465--1472,
  2011.

\bibitem{Kong2016HyperNet}
T.~Kong, A.~Yao, Y.~Chen, and F.~Sun.
\newblock Hypernet: Towards accurate region proposal generation and joint
  object detection.
\newblock In {\em Computer Vision and Pattern Recognition}, pages 845--853,
  2016.

\bibitem{L1998Gradient}
Y.~Lecun, L.~Bottou, Y.~Bengio, and P.~Haffner.
\newblock Gradient-based learning applied to document recognition.
\newblock {\em Proceedings of the IEEE}, 86(11):2278--2324, 1998.

\bibitem{Lifshitz2016Human}
I.~Lifshitz, E.~Fetaya, and S.~Ullman.
\newblock Human pose estimation using deep consensus voting.
\newblock In {\em European Conference on Computer Vision}, pages 246--260,
  2016.

\bibitem{lin2016fpn}
T.-Y. Lin, P.~Doll\'{a}r, R.~Girshick, K.~He, B.~Hariharan, and S.~Belongie.
\newblock Feature pyramid networks for object detection.
\newblock In {\em {CVPR}}, 2017.

\bibitem{Lin2014Microsoft}
T.~Y. Lin, M.~Maire, S.~Belongie, J.~Hays, P.~Perona, D.~Ramanan, P.~Dollár,
  and C.~L. Zitnick.
\newblock Microsoft coco: Common objects in context.
\newblock 8693:740--755, 2014.

\bibitem{2016arXiv161105424N}
A.~{Newell}, Z.~{Huang}, and J.~{Deng}.
\newblock {Associative Embedding: End-to-End Learning for Joint Detection and
  Grouping}.
\newblock {\em ArXiv e-prints}, Nov. 2016.

\bibitem{Newell2016Stacked}
A.~Newell, K.~Yang, and J.~Deng.
\newblock Stacked hourglass networks for human pose estimation.
\newblock In {\em European Conference on Computer Vision}, pages 483--499,
  2016.

\bibitem{2017arXiv170101779P}
G.~{Papandreou}, T.~{Zhu}, N.~{Kanazawa}, A.~{Toshev}, J.~{Tompson},
  C.~{Bregler}, and K.~{Murphy}.
\newblock {Towards Accurate Multi-person Pose Estimation in the Wild}.
\newblock {\em ArXiv e-prints}, Jan. 2017.

\bibitem{classic_4}
L.~Pishchulin, M.~Andriluka, P.~Gehler, and B.~Schiele.
\newblock Poselet conditioned pictorial structures.
\newblock In {\em Computer Vision and Pattern Recognition}, pages 588--595,
  2013.

\bibitem{Pishchulin2016DeepCut}
L.~Pishchulin, E.~Insafutdinov, S.~Tang, B.~Andres, M.~Andriluka, P.~Gehler,
  and B.~Schiele.
\newblock Deepcut: Joint subset partition and labeling for multi person pose
  estimation.
\newblock In {\em Computer Vision and Pattern Recognition}, pages 4929--4937,
  2016.

\bibitem{ren2015faster}
S.~Ren, K.~He, R.~Girshick, and J.~Sun.
\newblock Faster {R-CNN}: Towards real-time object detection with region
  proposal networks.
\newblock In {\em Neural Information Processing Systems ({NIPS})}, 2015.

\bibitem{ILSVRC15}
O.~Russakovsky, J.~Deng, H.~Su, J.~Krause, S.~Satheesh, S.~Ma, Z.~Huang,
  A.~Karpathy, A.~Khosla, M.~Bernstein, A.~C. Berg, and L.~Fei-Fei.
\newblock {ImageNet Large Scale Visual Recognition Challenge}.
\newblock {\em International Journal of Computer Vision (IJCV)},
  115(3):211--252, 2015.

\bibitem{classic_0}
B.~Sapp, C.~Jordan, and B.~Taskar.
\newblock Adaptive pose priors for pictorial structures.
\newblock In {\em Computer Vision and Pattern Recognition}, pages 422--429,
  2010.

\bibitem{classic_5}
B.~Sapp and B.~Taskar.
\newblock Modec: Multimodal decomposable models for human pose estimation.
\newblock In {\em Computer Vision and Pattern Recognition}, pages 3674--3681,
  2013.

\bibitem{ohem}
A.~Shrivastava, A.~Gupta, and R.~Girshick.
\newblock Training region-based object detectors with online hard example
  mining.
\newblock In {\em IEEE Conference on Computer Vision and Pattern Recognition},
  pages 761--769, 2016.

\bibitem{resinc}
C.~{Szegedy}, S.~{Ioffe}, V.~{Vanhoucke}, and A.~{Alemi}.
\newblock {Inception-v4, Inception-ResNet and the Impact of Residual
  Connections on Learning}.
\newblock {\em ArXiv e-prints}, Feb. 2016.

\bibitem{Tompson2014Joint}
J.~Tompson, A.~Jain, Y.~Lecun, and C.~Bregler.
\newblock Joint training of a convolutional network and a graphical model for
  human pose estimation.
\newblock {\em Eprint Arxiv}, pages 1799--1807, 2014.

\bibitem{Toshev2013DeepPose}
A.~Toshev and C.~Szegedy.
\newblock Deeppose: Human pose estimation via deep neural networks.
\newblock pages 1653--1660, 2013.

\bibitem{cpm}
S.-E. Wei, V.~Ramakrishna, T.~Kanade, and Y.~Sheikh.
\newblock Convolutional pose machines.
\newblock In {\em CVPR}, 2016.

\bibitem{Wei2016Convolutional}
S.~E. Wei, V.~Ramakrishna, T.~Kanade, and Y.~Sheikh.
\newblock Convolutional pose machines.
\newblock pages 4724--4732, 2016.

\bibitem{resnext}
S.~Xie, R.~Girshick, P.~Dollar, Z.~Tu, and K.~He.
\newblock Aggregated residual transformations for deep neural networks.
\newblock In {\em The IEEE Conference on Computer Vision and Pattern
  Recognition (CVPR)}, July 2017.

\bibitem{Yang_2017_ICCV}
W.~Yang, S.~Li, W.~Ouyang, H.~Li, and X.~Wang.
\newblock Learning feature pyramids for human pose estimation.
\newblock In {\em The IEEE International Conference on Computer Vision (ICCV)},
  Oct 2017.

\bibitem{Feature_Pyramids}
W.~Yang, S.~Li, W.~Ouyang, H.~Li, and X.~Wang.
\newblock Learning feature pyramids for human pose estimation.
\newblock In {\em The IEEE International Conference on Computer Vision (ICCV)},
  Oct 2017.

\bibitem{classic_1}
Y.~Yang and D.~Ramanan.
\newblock Articulated pose estimation with flexible mixtures-of-parts.
\newblock In {\em Computer Vision and Pattern Recognition}, pages 1385--1392,
  2011.

\end{thebibliography}
}
\end{document}